\documentclass[10pt,twocolumn]{article}

\usepackage[top=0.8in, bottom=0.8in, left=0.8in, right=0.8in]{geometry}

\usepackage{graphicx}
\usepackage{comment}
\usepackage{hyperref}

\DeclareGraphicsExtensions{.pdf,.png, .jpg, .jpeg}

\usepackage{mathtools} 
\usepackage{amsmath}   
\usepackage{amsfonts}  
\usepackage{bm}
\usepackage{verbatim}  

\usepackage{tabularx}
\usepackage{caption}
\usepackage{multirow}
\usepackage{booktabs}
\usepackage{flushend}

\newcolumntype{P}[1]{>{\centering\arraybackslash}p{#1}}

\begin{document}

\title{Automatic Sleep Stage Scoring with Single-Channel EEG Using Convolutional Neural Networks}

\author{Orestis~Tsinalis,
        Paul~M.~Matthews,
        Yike~Guo,
        and~Stefanos~Zafeiriou%
\thanks{O. Tsinalis, Y. Guo, and S. Zafeiriou are with the Department of Computing, Imperial College London, London, SW7 2AZ, United Kingdom (email: orestis.tsinalis10@imperial.ac.uk).}%
\thanks{P. M. Matthews is with the Division of Brain Sciences, Department of Medicine, Imperial College London.}%
}
\date{}

\maketitle

\begin{abstract}
We used convolutional neural networks (CNNs) for automatic sleep stage scoring based on single-channel electroencephalography (EEG) to learn task-specific filters for classification without using prior domain knowledge. We used an openly available dataset from 20 healthy young adults for evaluation and applied 20-fold cross-validation. We used class-balanced random sampling within the stochastic gradient descent (SGD) optimization of the CNN to avoid skewed performance in favor of the most represented sleep stages. We achieved high mean $F_1$-score (81\%, range 79--83\%), mean accuracy across individual sleep stages (82\%, range 80--84\%) and overall accuracy (74\%, range 71--76\%) over all subjects. By analyzing and visualizing the filters that our CNN learns, we found that rules learned by the filters correspond to sleep scoring criteria in the American Academy of Sleep Medicine (AASM) manual that human experts follow. Our method's performance is balanced across classes and our results are comparable to state-of-the-art methods with hand-engineered features. We show that, without using prior domain knowledge, a CNN can automatically learn to distinguish among different normal sleep stages.
\end{abstract}


\section{Introduction}
Convolutional neural networks (CNNs) are perhaps the most widely used technique in the deep learning class of machine learning algorithms \cite{lecun2015}. Their most important characteristic is that they learn task-specific filters without using any prior domain knowledge. CNNs have proven extremely effective in computer vision in areas such as object recognition, image segmentation and face recognition. The key to the success of CNNs has been \textit{end-to-end learning}, i.e. the integration of feature extraction and classification into a single algorithm using only the `raw' data (e.g. pixels, in the case of computer vision applications) as input. In biomedical engineering the adoption of CNNs has been uneven. On the one hand, the advances in CNNs for computer vision have been rapidly transferred in applications that are based on two-dimensional images---most notably in medical imaging. On the other hand, this has not been the case for biomedical applications which focus on classifying one-dimensional biosignals, such as electroencephalography (EEG) and electrocardiography (ECG). However, there has recently been a small but growing interest in using CNNs for biosignal-related problems \cite{ren2014, cecotti2014, kiranyaz2015, zhu2014}, including on the Kaggle platform \cite{kaggle2015} with EEG signals. In this paper we present a CNN architecture which we developed for automatic sleep stage scoring using a single channel of EEG.

Sleep is central to human health, and the health consequences of reduced sleep, abnormal sleep patterns or desynchronized circadian rhythms can be emotional, cognitive, or somatic \cite{wulff2010}. Associations between disruption of normal sleep patterns and neurodegenerative diseases are well recognized \cite{wulff2010}. According to the American Academy of Sleep Medicine (AASM) manual \cite{iber2007}, sleep is categorized into four stages. These are Rapid Eye Movement (stage R) sleep and 3 non-R stages, N1, N2 and N3. Formerly, stage N3 (also called Slow Wave Sleep, or SWS) was divided into two distinct stages, N3 and N4 \cite{recht1968}. To these a Wake (W) stage is added. These stages are defined by electrical activity recorded from sensors placed at different parts of the body. The totality of the signals that are recorded through these sensors is called a polysomnogram (PSG). The PSG includes an electroencephalogram (EEG), an electrooculogram (EOG), an electromyogram (EMG), and an electrocardiogram (ECG). After the PSG is recorded, it is divided into 30-second intervals, called \textit{epochs}. One or more experts then classify each epoch into one of the five stages (N1, N2, N3, R or W) by quantitatively and qualitatively examining the  signals of the PSG in the time and frequency domains. Sleep scoring is performed according to the Rechtschaffen and Kales sleep staging criteria \cite{recht1968}. In Table \ref{table:scoring_criteria} we reproduce the Rechtschaffen and Kales sleep staging criteria \cite{silber2007}, merging the criteria for N3 and N4 into a single stage (N3). Sleep stage scoring by human experts demands specialized training and thus can be expensive or difficult to access.

\begin{table*}
\centering
\footnotesize
\caption{The Rechtschaffen and Kales sleep staging criteria \cite{recht1968}, adapted from \cite{silber2007}.}
\label{table:scoring_criteria}
\begin{tabularx}{1\textwidth}{p{3cm} X}
\toprule
\rule{0pt}{0cm}
\textbf{Sleep Stage} & \textbf{Scoring Criteria} \\
\midrule
\rule{0pt}{0cm}
Non-REM 1 (N1) & $50\%$ of the epoch consists of relatively low voltage mixed (2-7 Hz) activity, and $<50\%$ of the epoch contains alpha (8-13 Hz) activity. Slow rolling eye movements lasting several seconds often seen in early N1.\\
\rule{0pt}{0.4cm}
Non-REM 2 (N2) & Appearance of sleep spindles and/or K complexes and $<20\%$ of the epoch may contain high voltage ($>75$ $\mu$V, $<2$ Hz) activity. Sleep spindles and K complexes each must last $>0.5$ seconds. \\
\rule{0pt}{0.4cm}
Non-REM 3 (N3) & $20\%-50\%$ (formerly N3) or $>50\%$ (formerly N4) of the epoch consists of high voltage ($>75$ $\mu$V), low frequency ($<2$ Hz) activity. \\
\rule{0pt}{0.4cm}
REM (R) & Relatively low voltage mixed (2-7 Hz) frequency EEG with episodic rapid eye movements and absent or reduced chin EMG activity. \\
\rule{0pt}{0.4cm}
Wake (W) & $> 50\%$ of the epoch consists of alpha (8-13 Hz) activity or low voltage, mixed (2-7 Hz) frequency activity. \\
\bottomrule
\end{tabularx}
\end{table*}

Recent research suggests that detection of sleep/circadian disruption could be a valuable marker of vulnerability and risk in the early stages of neurodegenerative diseases, such as Alzheimer's disease and Parkinson's disease, and that treatment of sleep pathologies can improve patient quality of life measures \cite{wulff2010}. Potential for widely accessible, longitudinal sleep monitoring would be ideal (for both medical research and medical practice). In this case an affordable, portable and unobtrusive sleep monitoring system for unsupervised at-home use would be needed. Wearable EEG is a strong candidate for such use. A core software component of such a system is a sleep scoring algorithm, which can reliably perform automatic sleep stage scoring given the patient's EEG signals.

\begin{table*}
\centering
\footnotesize
\caption{The transition rules summarised from the AASM sleep scoring manual\\ \cite[Chapter IV: Visual Rules for Adults, pp. 23--31]{iber2007}.}
\label{table:transition_rules}
\begin{tabularx}{1\textwidth}{p{2.6cm} p{4.1cm} p{2.2cm} X}
\toprule
\textbf{Sleep Stage Pair} & \textbf{Transition Pattern*} & \textbf{Rule} & \textbf{Differentiating Features} \\
\midrule
\multirow{4}{*}{N1-N2} & N1-\{N1,N2\} & 5.A.Note.1 & Arousal, K-complexes, sleep spindles \\
\cline{2-4}
\rule{0pt}{0.4cm} 
& \multirow{2}{*}{(N2-)N2-\{N1,N2\}(-N2)} & 5.B.1 & K-complexes, sleep spindles \\
&  & 5.C.1.b & Arousal, K-complexes, sleep spindles \\
\cline{2-4}
\rule{0pt}{0.4cm}
& N2-\{N1-N1,N2-N2\}-N2 & 5.C.1.c & Alpha, body movement, slow eye movement \\
\midrule
\multirow{4}{*}{N1-R} & \multirow{3}{*}{R-R-\{N1,R\}-N2} & 7.B & Chin EMG tone \\
& & 7.C.1.b & Chin EMG tone \\
& & 7.C.1.c & Chin EMG tone, arousal, slow eye movement \\
\cline{2-4}
\rule{0pt}{0.4cm}
& R-\{N1-N1-N1,R-R-R\} & 7.C.1.d & Alpha, body movement, slow eye movement \\
\midrule
\multirow{4}{*}{N2-R} & R-R-\{N2,R\}-N2 & 7.C.1.e &  Sleep spindles \\
\cline{2-4}
\rule{0pt}{0.4cm}
& \multirow{3}{*}{(N2-)N2-\{N2,R\}-R(-R)} & 7.D.1 & Chin EMG tone \\
& & 7.D.2 & Chin EMG tone, K-complexes, sleep spindles \\
& & 7.D.3 & K-complexes, sleep spindles \\
\bottomrule
\multicolumn{4}{p{17.1cm}}{\rule{0pt}{0.3cm}*Curly braces indicate choice between the stages or stage progressions in the set, and parentheses indicate optional epochs.}\\
\end{tabularx}
\end{table*}

In this study, we present and evaluate a novel CNN architecture for automatic sleep stage scoring using a single channel of EEG. We compared the performance of CNN with our previous study \cite{tsinalis2015}, in which we hand-engineered the features for classification. In that study we used the Fpz-Cz electrode and time-frequency analysis-based feature extraction fine-tuned to capture sleep stage-specific signal features using Morlet wavelets (see for example, Chapters 12 and 13, pp. 141--174 in \cite{cohen2014}), with stacked sparse autoencoders \cite{bengio2006, bengio2009} as the classification algorithm. In that work we had achieved state-of-the-art results, compared to the existing studies in \cite{fraiwan2012, liang2012, berthomier2007}, mitigated skewed sleep scoring performance in favor of the most represented sleep stages, and addressed the problem of misclassification errors due to class imbalance in the training data while significantly improving worst-stage classification. We will use this work \cite{tsinalis2015} for comparison with the results from the new approach presented here.

\section{Materials and Methods}
\subsection{Data}
The dataset that we used to evaluate our method is a publicly available sleep PSG dataset \cite{kemp2000} from the PhysioNet repository \cite{goldberger2000} that can be downloaded from \cite{physionetdb}. The data was collected from electrodes Fpz-Cz and Pz-Oz. The sleep stages were scored according to the Rechtschaffen and Kales guidelines \cite{recht1968}. The epochs of each recording were scored by a single expert (6 experts in total). The sleep stages that are scored in this dataset are Wake (W), REM (R), non-R stages 1--4 (N1, N2, N3, N4), Movement and Not Scored. For our study, we removed the very small number of Movement and Not Scored epochs (Not Scored epochs were at the start or end of each recording), and also merged the N3 and N4 stages into a single N3 stage, as is currently the recommended by the American Academy of Sleep Medicine (AASM) \cite{iber2007, silber2007}. There were 61 movement epochs in our data in total, and only 17/39 recordings had movement artifacts. The maximum number of movement epochs per recording was 12. The rationale behind the decision of removing the movement epochs was based on two facts. First, these epochs had not been scored by the human expert as belonging to any of the 5 sleep stages, as it is recommended in the current AASM manual \cite[p. 31]{iber2007}. Second, their number was so small that they could not be used as a separate `movement class' for learning. The public dataset includes 20 healthy subjects, 10 male and 10 female, aged 25--34 years. There are two approximately 20-hour recordings per subject, apart from a single subject for whom there is only a single recording. To evaluate our method we used the in-bed part of the recording. The sampling rate is 100 Hz and the epoch duration is 30 seconds. 

\subsection{Convolutional neural network architecture}
A convolutional neural network (CNN) is composed of successive convolutional (filtering) and pooling (subsampling) layers with a form of nonlinearity applied before or after pooling, potentially followed by one or more fully-connected layers. In classification problems, like sleep scoring, the last layer of a CNN is commonly a softmax (multinomial logistic regression) layer. CNNs are trained using iterative optimization with the backpropagation algorithm. The most common optimization method in the literature is stochastic gradient descent (SGD).

In our CNN architecture we are using the raw EEG signal without preprocessing as the input. Using raw input (usually with some preprocessing) in CNN architectures is the norm in applications of deep learning in computer vision. In classification problems with one-dimensional (1D) signals CNNs can also be applied to a precomputed spectrogram or other time-frequency decomposition of the signal, so that the input to the CNN is a two-dimensional (2D) stack of frequency-specific activity over time. Characteristic examples of this approach can be found in recent work in signal processing for speech and acoustics \cite{sainath2013, dielman2014, huang2015, zhang2015}. When the spectrogram is used as input it can be treated as a 2D image. Recently, there has been also growing interest in applying CNNs to raw 1D signals. Again, there are characteristic examples from speech and acoustics in \cite{dielman2014, palaz2014, swietojanski2014, palaz2015, hoshen2015}.

\begin{figure*}
    \centering
    \includegraphics[width=\textwidth]{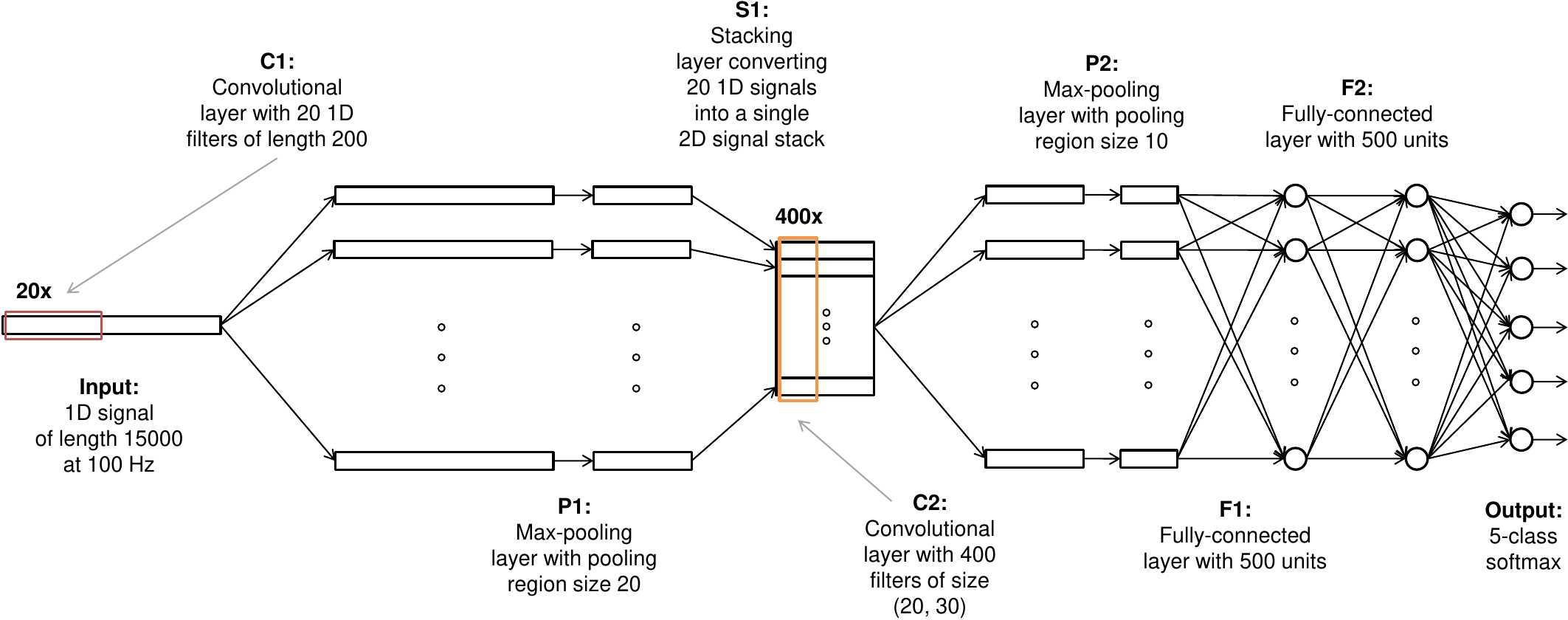}
    \caption{CNN architecture}
    \label{fig:cnn_architecture}
\end{figure*}

Our CNN architecture, shown in Figure \ref{fig:cnn_architecture}, comprises two pairs of convolutional and pooling layers (C1-P1 and C2-P2), two fully-connected layers (F1 and F2), and a softmax layer. Between layer P1 and layer C2, we include a `stacking layer', S1. As shown in Table \ref{table:cnn_architecture}, layer C1 contains 20 filters, so that the output of layer C1 is 20 filtered versions of the original input signal. These filtered signals are then subsampled in layer P1. The stacking layer rearranges the output of the layer P1, so that instead of 20 distinct signals the input to the next convolutional layer C2 is a 2D stack of filtered and subsampled signals. As shown in Figure \ref{fig:cnn_architecture} and Table \ref{table:cnn_architecture}, the filters in layer C2 are 2D filters. The height of the layer C2 filters is 20, same as the height of the stack. The purpose of these 2D filters is to capture relationships across the filtered signals produced by filtering the original signal in layer C1, across a specific time window.

With this CNN architecture we attempt to combine a CNN architecture using raw signals \cite{dielman2014, palaz2014, swietojanski2014, palaz2015, hoshen2015} with the idea of using a 2D stack of frequency-specific activity over time \cite{sainath2013, dielman2014, huang2015, zhang2015}. In a standard CNN architecture layer C2 would have the same structure as layer C1, with a number of 1D filters applied to each of layer P1 outputs. The most common way to combine information across the layer P2 outputs is by adding up the filtered signals of layer C2 across layer P2 outputs by filter index. While this has an effect similar to the stacking layer, we think that explicitly stacking the outputs of layer P2 makes clear the correspondence between CNN methods and hand-engineered feature methodologies.

The cost function for the training of our CNN architecture was the softmax with L2-regularization. We applied the rectified linear unit (ReLU) nonlinearity after convolution and before pooling. The hyperparameters of a CNN are: the number and types of layers, the size of the filters for convolution and the convolution stride for each convolutional layer, the pooling region size and the pooling stride for each pooling layer, and the number of units for each fully-connected layer. We summarize the selected hyperparameters for our CNN architecture in Table \ref{table:cnn_architecture}.
 
The classes (sleep stages) in our dataset, as in any PSG dataset, were not balanced, i.e. there were many more epochs for some stages (particularly N2) than others (particularly W and N1). In such a situation, if all the data is used as is, it is highly likely that a classifier will exhibit skewed performance favoring the most represented classes, unless the least represented classes are very distinct from the other classes. In order to resolve the issues stemming from imbalanced classes, in our previous work \cite{tsinalis2015} we employed class-balanced random sampling with an ensemble of 20 classifiers, each one being trained on a different balanced sample of the data. This is not an efficient way for class-balancing with CNNs, as training even a single CNN is very time-consuming. The strategy that we followed in the current paper was different. At each epoch of SGD we used a different class-balanced batch for the optimization.

As shown in Table \ref{table:transition_rules} the scoring of a particular epoch can depend on the characteristics of the preceding or succeeding epochs, for the sleep stage pairs N1-N2, N1-R, and N2-R. Therefore, we chose the input data to our CNN to be the signal of the current epoch to be classified together with the signals of the preceding two and succeeding two epochs, as a single, continuous signal, starting from the earliest epoch, with the current epoch in the middle. At the sampling rate of 100 Hz this gives an input size of 15,000 timepoints.

We implemented the different CNN architectures using the Python libraries Lasagne (\url{https://github.com/Lasagne/Lasagne}) and Theano (\url{https://github.com/Theano/Theano}).

\begin{table*}
\centering
\caption{CNN architecture}
{
\label{table:cnn_architecture}
\begin{tabularx}{1\textwidth}{X P{2.6cm} P{1.8cm} P{1.8cm} P{1.8cm} P{1.8cm} P{2.3cm}}
\toprule
\rule{0pt}{0cm}
\textbf{Layer} & \textbf{Layer Type} & \textbf{\# Units} & \textbf{Unit Type} & \textbf{Size} & \textbf{Stride} & \textbf{Output Size} \\
\midrule
\rule{0pt}{0cm}
Input & & & & & & (1, 1, 15000)\\
\rule{0pt}{0.3cm}
C1 & convolutional & 20 & ReLU & (1, 200) & (1, 1) & (20, 1, 14801) \\
\rule{0pt}{0.3cm}
P1 & max-pooling & & & (1, 20) & (1, 10) & (20, 1, 1479) \\
\rule{0pt}{0.3cm}
S1 & stacking & & & & & (1, 20, 1479) \\
\rule{0pt}{0.3cm}
C2 & convolutional & 400 & ReLU & (20, 30) & (1, 1) & (400, 1, 1450) \\
\rule{0pt}{0.3cm}
P2 & max-pooling & & & (1, 10) & (1, 2) & (400, 1, 721) \\
\rule{0pt}{0.3cm}
F1 & fully-connected & 500 & ReLU & & & 500 \\
\rule{0pt}{0.3cm}
F2 & fully-connected & 500 & ReLU & & & 500 \\
\rule{0pt}{0.3cm}
Output & softmax & 5 & logistic & & & 5 \\
\bottomrule
\end{tabularx}
}
\end{table*}

\subsection{Evaluation}
To evaluate the generalizability of the algorithms, we obtained results using 20-fold cross-validation as in \cite{tsinalis2015}. Specifically, in each fold we use the recordings of a single subject for testing and the recordings of the remaining 19 subjects for training and validation. For each fold we used the recordings from 4 randomly selected subjects as validation data and the recordings from the remaining 15 subjects for training. The classification performance in the validation data was used for choosing the hyperparameters and as a stopping criterion for training to avoid overfitting to the training data. 

All scoring performance metrics were derived from the confusion matrix. Using a `raw' confusion matrix in the presence of imbalanced classes implicitly assumes that the relative importance of correctly detecting a class is directly proportional to its frequency of occurrence. This is not desirable for sleep staging. What we need to mitigate the negative effects of imbalanced classes on classification performance measurement is effectively a normalized or `class-balanced' confusion matrix that places equal weight into each class.

The metrics we computed were precision, sensitivity, $F_1$-score, per-stage accuracy, and overall accuracy. The $F_1$-score is the harmonic mean of precision and sensitivity and is a more comprehensive performance measure than precision and sensitivity by themselves. The reason is that precision and sensitivity can each be improved at the expense of the other. All the metrics apart from overall accuracy are binary. However, in our case we have 5 classes. Therefore, after we performed the classification and computed the normalized confusion matrix, we converted our problem into 5 binary classification problems each time considering a single class as the `positive' class and all other classes combined as a single `negative' class (\textit{one-vs-all} classification).  

We report the evaluation metrics across all folds. Specifically, we report their mean value across all 5 sleep stages and their value for the most misclassified sleep stage, which provides information about the robustness of the method across sleep stages. We tested our method with the Fpz-Cz electrode, with which we had achieved better performance in \cite{tsinalis2015}.

We calculated 95\% confidence intervals for each of the performance metrics by bootstrapping using 1000 bootstrap samples across the confusion matrices of the 39 recordings. For each bootstrap sample we sampled the recording indexes (from 1 to 39) with replacement and then added up the confusion matrices of the selected recordings. We then calculated each evaluation metric for each bootstrap sample. We report the mean value of each metric across the bootstrap samples, and the values that define the range of the 95\% confidence interval per metric, i.e. the value of the metric in the 26th and 975th position of the ordered bootstrap sample metric values. 

To further evaluate the generalizability of our method, we performed two tests on our results to assess the correlation between scoring performance and (1) a measure of the sleep quality of each recording, and (2) the percentage of transitional epochs in each recording. Robust scoring performance across sleep quality and temporal sleep variability, can be seen as further indicators of the generalizability of an automatic sleep stage scoring algorithm. The reason is that low sleep quality and high sleep stage variability across the hypnogram are prevalent in sleep pathologies (see, for example, \cite{norman2000}).

We measured sleep quality with a widely-used index, called \textit{sleep efficiency}. Sleep efficiency is defined as the percentage of the total time in bed that a subject was asleep \cite[p. 226]{spriggs2014}. Our data contain a `lights out' indicator, which signifies the start of the time in bed. We identified the sleep onset as the first non-W epoch that occurred after lights were out. We identified the end of sleep as the last non-W epoch after sleep onset, as our dataset does not contain a `lights on' indicator. The number of epochs between the start of time in bed and the end of sleep was the total time in bed, within which we counted the non-W epochs; this was the total time asleep. We defined transitional epochs as those whose preceding or succeeding epochs were of a different sleep stage than them. We computed their percentage with respect to the total time in bed. In our experiments we computed the $R^2$ and its associated p-value between sleep efficiency and scoring performance, and between percentage of transitional epochs and scoring performance.

We compared our new, CNN results with our previous work \cite{tsinalis2015}, as well as with those from a CNN architecture that uses the same Morlet wavelets as in \cite{tsinalis2015} to produce a time-frequency stack that is fed to the CNN from the second convolutional layer C2 onwards.

\subsection{CNN filter analysis and visualization}
Apart from performance evaluation an additional type of evaluation is required when using CNNs, in our view. As the filters in CNNs are automatically learned from the training data, we need to evaluate whether the filters learned in different folds (i.e. using different training data) are similar across folds. We analyzed and compared the learned filters from the first convolutional layer of the CNN from each of the 20 different folds. For all of the architectures layer C1 has 20 filters. We extracted the frequency content of the filters by computing the power at different frequency bands using the Fourier transform.

We then fed the testing data for that fold to the CNN. We extracted the features produced by each filter per training example for the middle segment of the signal (the current epoch). Each feature is a signal which represents the presence of the filter over time. We computed the power of the feature signal for each testing example, and then took the mean power across all testing examples of each true (not predicted) class. Some filters have naturally lower power, because they correspond to patterns localized in time and not in continuous activity as shown in the scoring criteria in Table \ref{table:scoring_criteria}.

We observed that certain sleep stages produce higher filter activations across all filters in general. To account for those differences, we normalized (to unit length) the power first by sleep stage across filters, and then by filter across sleep stages. Similar filters learned in each fold are generally not at the same index. For easier visual inspection of the results we ordered the filters by the sleep stage for which they have the greatest mean activation.

Finally, we qualitatively compared the learned filters with the guidelines in the AASM sleep scoring manual. To do so we also compared the filters and activation patterns per filter per sleep stage with the frequency content and activation patterns of the hand-engineered Morlet wavelets we used in \cite{tsinalis2015}.

\section{Results}

\subsection{Sleep stage scoring performance}
As we show in the normalized confusion matrix in Table \ref{table:confusion_matrix_FpzCz}, the most correctly classified sleep stage was N3 with around 90\% of stage N3 epochs correctly classified. Stages R and N2 follow with around 75\% of epochs correctly classified for each stage. Stage W has around 70\% of epochs correctly classified. The most misclassified sleep stage was N1 with 60\% of stage N1 epochs correctly classified. Most misclassifications occurred between the pairs N1-W and N1-R (about 15\%), followed by pairs N1-N2, N2-R and N2-N3 (about 8\%), and R-W and N2-W (about 5\%). The remaining pairs, N1-N3, N3-R and N3-W have misclassification rates close to zero.

\begin{table*}
\centering
\footnotesize
\caption{Confusion matrix from cross-validation using the Fpz-Cz electrode.}
{
\label{table:confusion_matrix_FpzCz}
\begin{tabularx}{1\textwidth}{X | P{2.4cm} P{2.4cm} P{2.4cm} P{2.4cm} P{2.4cm}}
\toprule
 & \textbf{N1} & \textbf{N2} & \textbf{N3} & \textbf{R} & \textbf{W} \\
 & \textbf{(algorithm)} & \textbf{(algorithm)} & \textbf{(algorithm)} & \textbf{(algorithm)} & \textbf{(algorithm)} \\
\midrule
\rule{0pt}{0cm}
\textbf{N1 (expert)} & \textbf{1657} (60\%) & \textbf{259} (9\%)  & \textbf{9} (0\%) & \textbf{427} (15\%) & \textbf{410} (15\%) \\
\rule{0pt}{0.3cm}
\textbf{N2 (expert)} & \textbf{1534} (9\%) & \textbf{12858} (73\%) & \textbf{1263} (7\%) & \textbf{1257} (7\%) & \textbf{666} (4\%) \\
\rule{0pt}{0.3cm}
\textbf{N3 (expert)} & \textbf{9} (0\%) & \textbf{399} (7\%) & \textbf{5097} (91\%) & \textbf{1} (0\%) & \textbf{85} (2\%) \\
\rule{0pt}{0.3cm}
\textbf{R (expert)} & \textbf{1019} (13\%) & \textbf{643} (8\%) & \textbf{3} (0\%) & \textbf{5686} (74\%) & \textbf{360} (5\%) \\
\rule{0pt}{0.3cm}
\textbf{W (expert)} & \textbf{605} (18\%) & \textbf{171} (5\%) & \textbf{47} (1\%) & \textbf{175} (5\%) & \textbf{2382} (70\%) \\
\bottomrule
\multicolumn{6}{P{17.1cm}}{\rule{0pt}{0.3cm}This confusion matrix is the sum of the confusion matrices from each fold. The numbers in bold are numbers of epochs. The numbers in parentheses are the percentage of epochs that belong to the class classified by the expert (rows) that were classified by our algorithm as belonging to the class indicated by the columns.}\\
\end{tabularx}
}
\end{table*}

The percentage of false negatives with respect to each stage (non-diagonal elements in each row) per pair of stages was approximately balanced between the stages in the pair. An exception is the pair N1-W, which appears slightly skewed (3\% difference) in favor of stage N1. Effectively the upper and lower triangle of the confusion matrix are close to being mirror images of each other. This is a strong indication that the misclassification errors due to class imbalance have been mitigated.

As we show in Table \ref{table:comparison}, our method has high mean $F_1$-score (79\%, range 81--83\%), mean accuracy across individual sleep stages (80\%, range 82--84\%) and overall accuracy (74\%, range 71--76\%) over all subjects. From the scoring performance metrics results in Table \ref{table:comparison} we observe that our method has slightly worse performance than our previous work in \cite{tsinalis2015}. We should note though that the 95\% confidence intervals overlap for the majority of the metrics (worst-stage precision, mean and worst-stage sensitivity, mean and worst-stage $F_1$-score, and worst-stage and overall accuracy), and are otherwise nearly overlapping for the remaining metrics (mean precision and mean accuracy).

\begin{figure}
    \centering
    \includegraphics[width=0.45\textwidth]{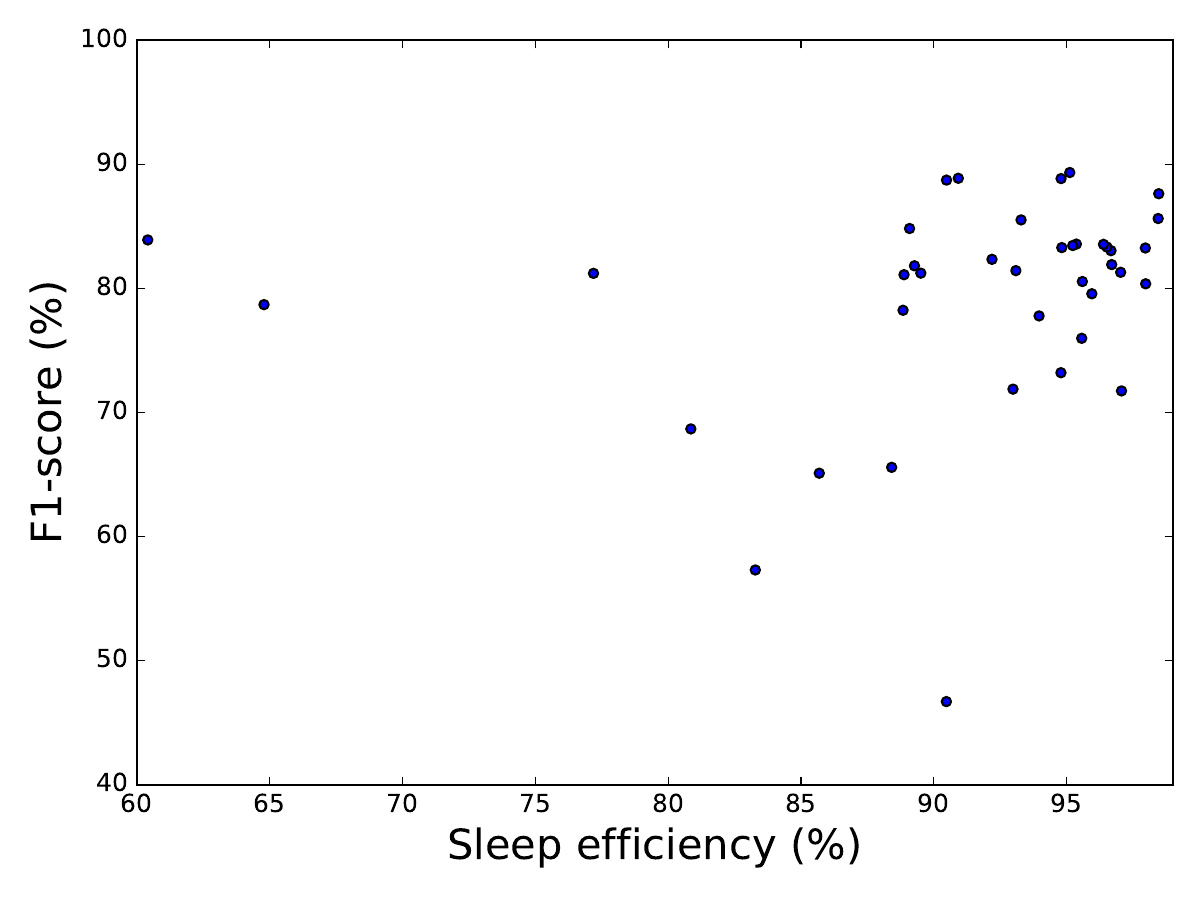}
    \caption{$F_1$-score as a function of sleep efficiency.}
    \label{fig:f1_efficiency}
\end{figure}

\begin{figure}
    \centering
    \includegraphics[width=0.45\textwidth]{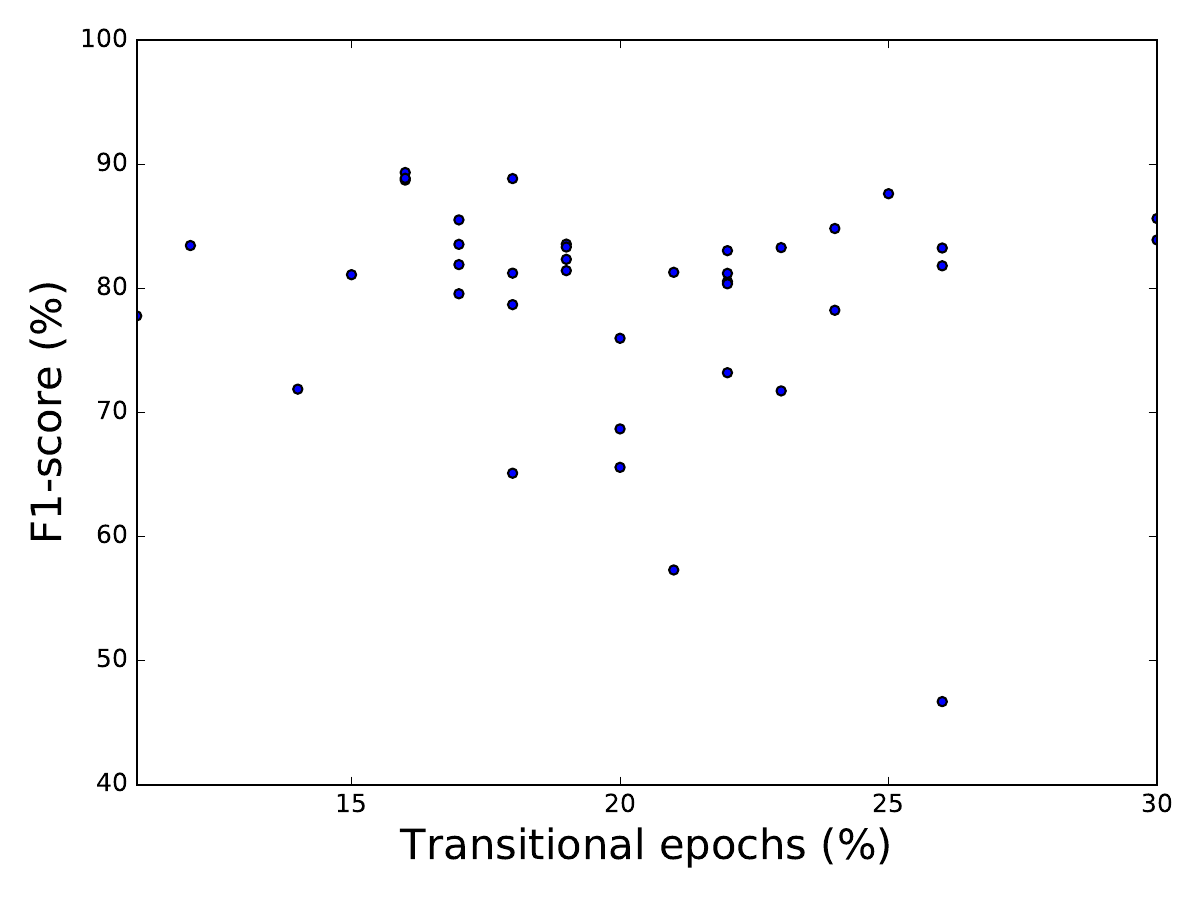}
    \caption{$F_1$-score as a function of transitional epochs.}
    \label{fig:f1_transitional}
\end{figure}

We also assessed the independence of the scoring performance (for $F_1$-score and overall accuracy) of our method across recordings relative to sleep efficiency and the percentage of transitional epochs per recording (Table \ref{table:correlation}). The p-values of the regression coefficients are all above 0.25. The $R^2$ is already negligible ($<0.05$) in all cases. For clarity, we present the data for these tests graphically for the $F_1$-score results in Figures \ref{fig:f1_efficiency} and \ref{fig:f1_transitional}. Our dataset contained 10 recordings with sleep efficiency below 90\% (in the range 60-89\%), which is the threshold recommended in \cite[p. 7]{spriggs2014} for young adults. The percentage of transitional epochs ranged from 10-30\% across recordings.

\begin{figure*}
    \centering
    \includegraphics[width=0.8\textwidth]{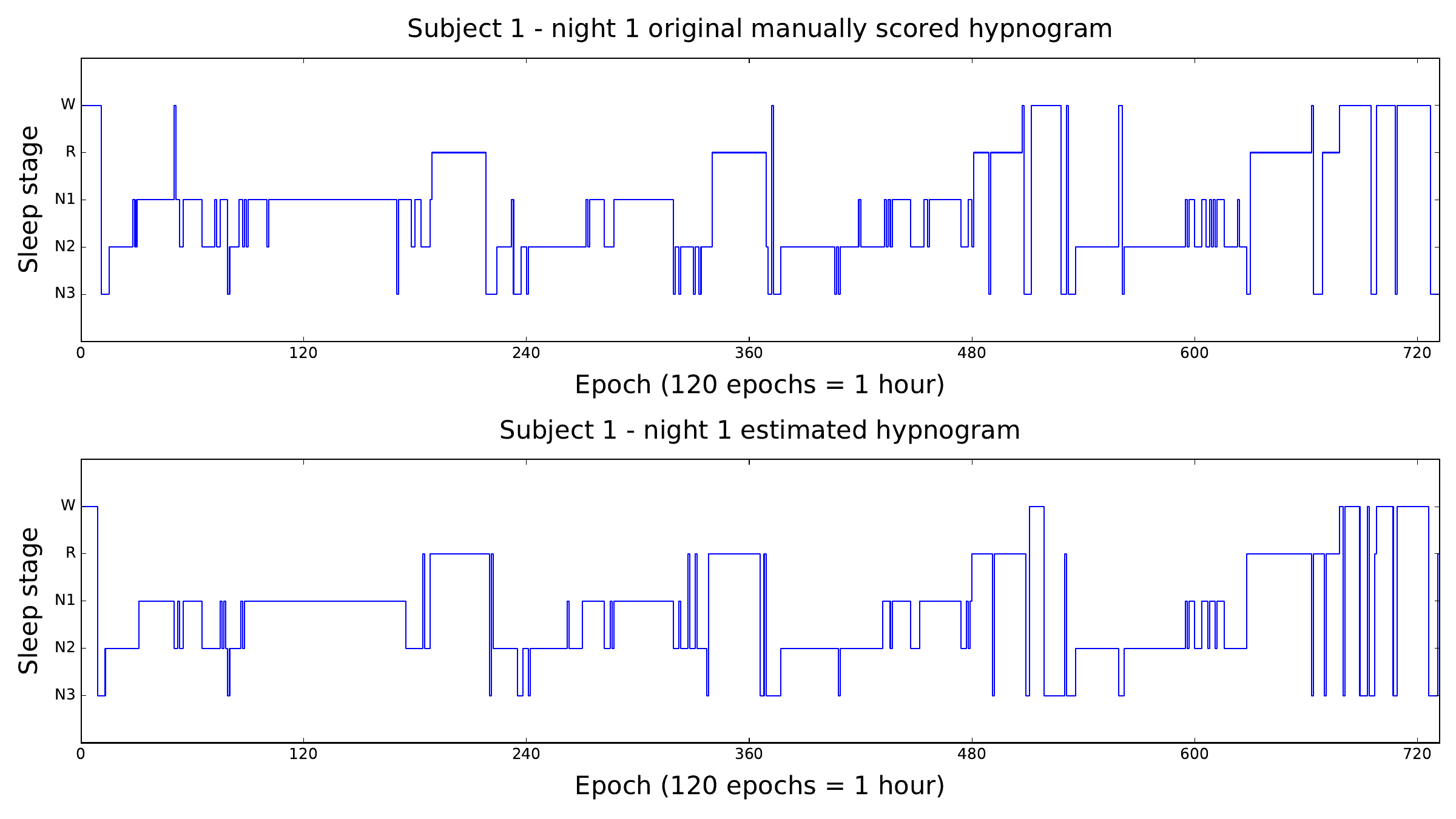}
    \caption{The original manually scored hypnogram (top) and the estimated hypnogram\\using our algorithm (bottom) for the first night of subject 1.}
    \label{fig:hypnograms}
\end{figure*}

Finally, in Figure \ref{fig:hypnograms} we present an original manually scored hypnogram and its corresponding estimated sleep hypnogram using our algorithm for a single PSG for which the overall $F_1$-score was approximately equal to the mean $F_1$-score across the entire dataset.

\begin{table*}
\centering
\footnotesize
\caption{Comparison between our CNN method and our previous state-of-the-art results with \\hand-engineered features \cite{tsinalis2015} on the same data set across the five scoring performance metrics\\(precision, sensitivity, $F_1$-score, per-stage accuracy, and overall accuracy).}
{
\label{table:comparison}
\begin{tabularx}{1\textwidth}{X P{1.2cm} P{1.2cm} P{1.2cm} P{1.2cm} P{1.2cm} P{1.2cm} P{1.2cm} P{1.2cm} P{1.2cm}}
\toprule
& \multicolumn{9}{c}{\textbf{Scoring performance metrics}} \\
\cmidrule{2-10}
& \multicolumn{2}{c}{\textbf{Precision}} & \multicolumn{2}{c}{\textbf{Sensitivity}} & \multicolumn{2}{c}{\textbf{$F_1$-score}} & \multicolumn{3}{c}{\textbf{Accuracy}}
\\
\midrule
\textbf{Study} & Mean & Worst & Mean & Worst & Mean & Worst & Mean & Worst & Overall \\
\midrule
& (92) & (86) & (75) & (55) & (82) & (68) & (84) & (74) & (75) \\
\cite{tsinalis2015} & \textbf{93} & \textbf{88} & \textbf{78} & \textbf{60} & \textbf{84} & \textbf{71} & \textbf{86} & \textbf{76} & \textbf{78} \\
& (94) & (90) & (80) & (65) & (86) & (75) & (88) & (78) & (80) \\
\midrule
CNN with & (90) & (82) & (71) & (48) & (79) & (61) & (80) & (67) & (71) \\
Morlet & \textbf{91} & \textbf{85} & \textbf{73} & \textbf{52} & \textbf{81} & \textbf{64} & \textbf{81} & \textbf{69} & \textbf{73} \\
wavelets & (92) & (87) & (75) & (56) & (83) & (68) & (83) & (72) & (75) \\
\midrule
& (90) & (84) & (71) & (53) & (79) & (66) & (80) & (70) & (71) \\
CNN & \textbf{91} & \textbf{86} & \textbf{74} & \textbf{60} & \textbf{81} & \textbf{70} & \textbf{82} & \textbf{73} & \textbf{74} \\
& (92) & (88) & (76) & (66) & (83) & (75) & (84) & (76) & (76) \\
\bottomrule
\multicolumn{10}{P{17.1cm}}{\rule{0pt}{0.3cm}For the binary metrics, we report the mean performance (over all five sleep stages) as well as the worst performance (in the most misclassified sleep stage, always stage N1). We present the results for our method using the Fpz-Cz electrode with 20-fold cross-validation. The numbers in parentheses are the bootstrap 95\% confidence interval bounds for the mean performance across subjects. The numbers in bold are the mean metrics values from bootstrap.}\\
\end{tabularx}
}
\end{table*}

\begin{table*}
\centering
\footnotesize
\caption{$R^2$ between sleep efficiency and percentage of transitional epochs,\\and scoring performance ($F_1$-score and overall accuracy).}
\label{table:correlation}
\begin{tabularx}{1\textwidth}{l X X X X}
\toprule
& \multicolumn{4}{c}{\textbf{Recording parameters}} \\
\cmidrule{2-5}
& \multicolumn{2}{c}{\textbf{Sleep efficiency}} & \multicolumn{2}{c}{\textbf{Percentage of transitional epochs}} \\
\textbf{Metric} & $R^2$ & p-value & $R^2$ & p-value \\
\midrule
$F_1$-score & 0.04 & 0.25 & 0.01 & 0.50 \\
Overall accuracy & 0.03 & 0.30 & 0.01 & 0.55 \\
\bottomrule
\end{tabularx}
\end{table*}

\subsection{CNN filter analysis and visualization}
\begin{figure*}
    \centering
    \includegraphics[width=\textwidth]{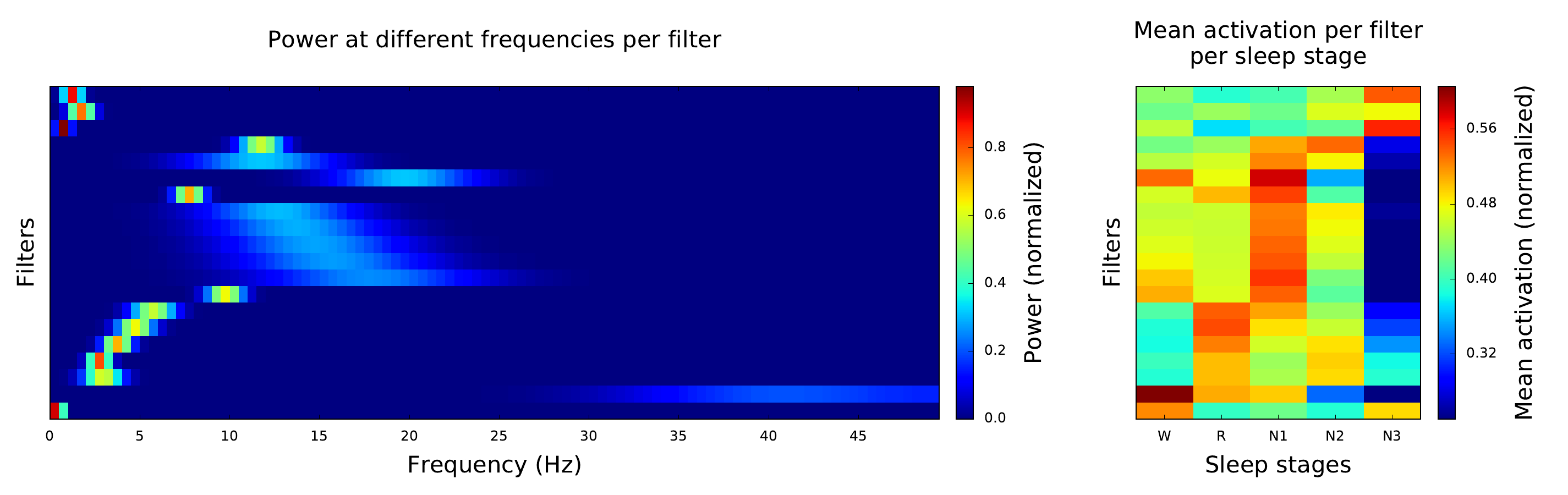}
    \caption{Filter visualization for the hand-engineered filters from \cite{tsinalis2015}.}
    \label{fig:morlet_filters}
\end{figure*}

We computed the frequency content and mean activation per sleep stage for the hand-engineered Morlet wavelet filters in \cite{tsinalis2015} as a reference. This visualization is shown in Figure \ref{fig:morlet_filters}. In Figure \ref{fig:visualization} we show the filter visualization for 5 folds of the cross-validation. This allows us to observe patterns of similarity between the filters learned using different subsets of subjects for training.

Our general observation is that the filters learned by the CNNs at different folds exhibit certain high-level similarities which are consistent across folds. We summarize the filter-sleep stage associations that are more prevalent in the visualization in Figure \ref{fig:visualization} (showing 5 of the folds), and are replicable across all folds. 

We observed that filters with highest power at 1-1.5 Hz, usually combined with 12.5-14 Hz are associated with highest activation in stage N3 epochs. Filters with highest power at 13-14.5 Hz, usually combined with 2-4 Hz, are associated with highest activation in stage N2 epochs. High power below 1 Hz filters are associated with highest activation in stage W epochs. Filters with highest power in frequencies 2-5 Hz mostly combined with 14 Hz are associated with highest activation in stage R epochs. It also is worth mentioning that the 2-5/14 Hz filters associated with stage R do not contain frequencies from 20-50 Hz. Stage N1 is commonly associated in the majority of folds with filters combining frequencies of 7 Hz and 9 Hz (but not 8 Hz), and always contain frequencies from 20-50 Hz. A common characteristic of all the CNN filters across folds is the absence of filters with frequencies from 10.5-12 Hz and from 15-16.5 Hz. 

\section{Discussion}
In Table \ref{table:comparison} we compare the performance of our previously published method with hand-engineered features and stacked sparse autoencoders \cite{tsinalis2015} (SAE model), our proposed CNN model, and an `intermediate' model which is using the hand-engineered Morlet wavelets of \cite{tsinalis2015} (shown in Figure \ref{fig:morlet_filters}) as the first fixed (i.e. \textit{untrainable}) layer of the CNN (M-CNN model) shown in Figure \ref{fig:cnn_architecture}. We should note that the architecture used for the M-CNN model was not optimized for the fixed filters, but is exactly the same as the CNN model, to allow us to assess the effect that fixing the filters in the first layer of our CNN model has.

The overall picture that arises from inspecting Table \ref{table:comparison} is that the SAE model outperforms the CNN model, which, in turn outperforms the M-CNN model. Worst-stage performance over all metrics is much closer between the SAE and the CNN model, although the SAE model is 3-4\% better in mean performance, which is almost identical between the CNN and the M-CNN model. However, for the M-CNN model, worst-stage performance is much lower than either the SAE or the CNN model. However, we observe that the 95\% confidence intervals across subjects overlap across the three models, across almost all of the metrics (the two exceptions are mean and worst-stage accuracy between the SAE and the M-CNN model). This indicates that the differences in performance across subjects are not statistically significant overall.

From the results in Table \ref{table:comparison} we observe two broad points. The first is that hand-engineering of features based on the AASM manual (SAE model) may have better performance than automatic filter learning (CNN model), although the difference based on the data set we used does not appear to be statistically significant. Using a larger data set could help clarify any differences in performance between the two models. In general, we expect that a larger data set would be beneficial for the performance of the CNN model, as CNN models can be difficult to train effectively with smaller data sets. The second point from Table \ref{table:comparison} is that using a fixed set of filters for the first CNN layer (M-CNN model) achieves worse performance than an end-to-end CNN (CNN model). However, the differences between the two models do not appear to be statistically significant.

Similarly to our previous work \cite{tsinalis2015} the CNN model exhibits balanced sleep scoring performance across sleep stages. The majority of misclassification errors is likely due especially to EOG and EMG-related patterns that are important in distinguishing between certain sleep stage pairs (see Tables \ref{table:scoring_criteria} and \ref{table:transition_rules}), which are difficult to capture through the single channel of EEG. We experimented with a number of filters larger than 20, but our results did not improve, and, in some cases, deteriorated. This corroborates our hypothesis that remaining misclassification errors may arise from not being able to capture patterns from other modalities of the PSG. 

Although we recognize that our dataset does not contain a very large number of recordings of bad sleep quality, we found no statistically significant correlation between sleep efficiency and mean scoring performance (see Table \ref{table:correlation} and Figure \ref{fig:f1_efficiency}). Similarly, there was not a significant correlation between the percentage of transitional epochs (which are by definition more ambiguous) and mean sleep scoring performance (see Table \ref{table:correlation} and Figure \ref{fig:f1_transitional}).

\begin{figure*}
    \centering
    \includegraphics[width=\textwidth]{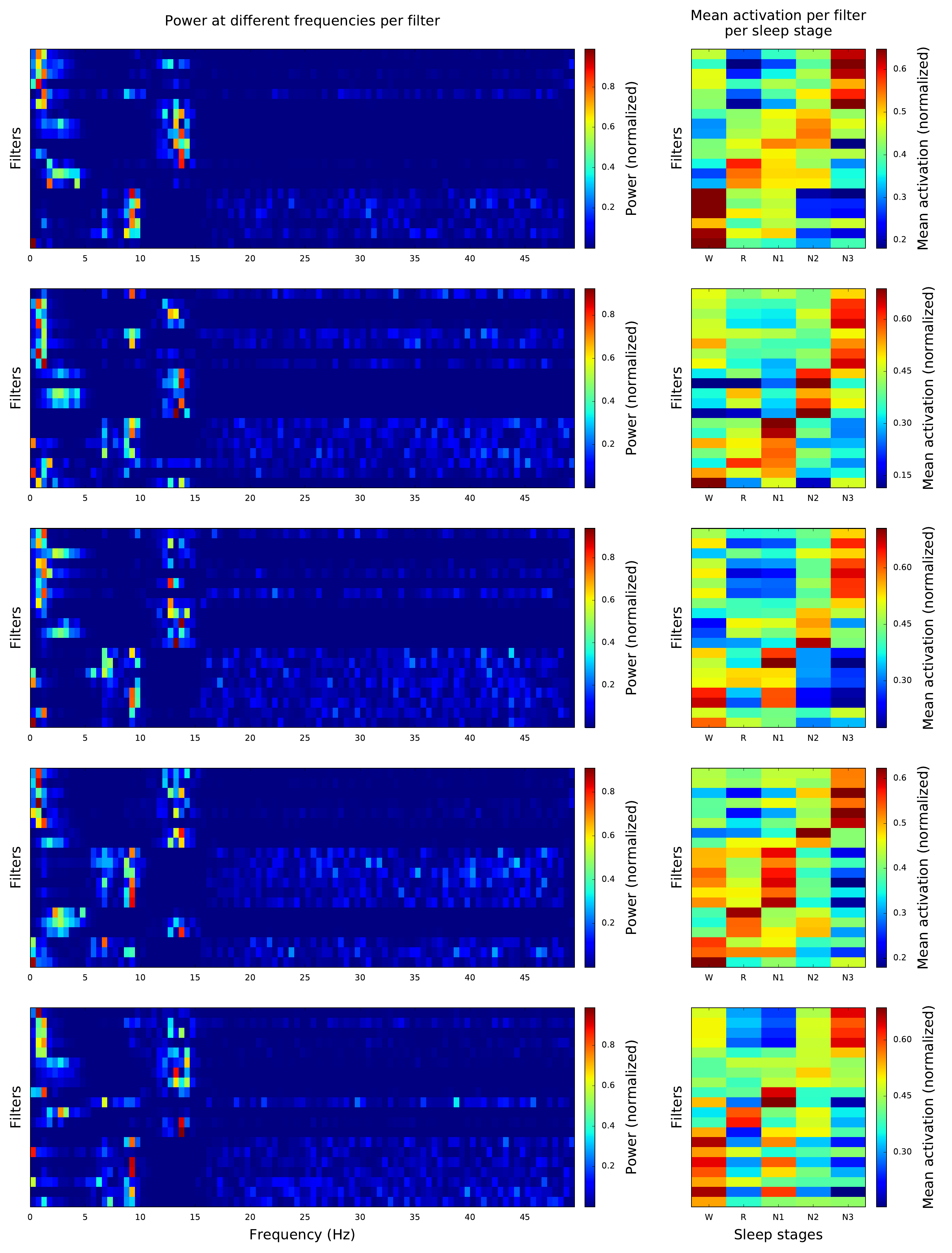}
    \caption{Filter visualization for folds 1, 5, 6, 8 and 19.}
    \label{fig:visualization}
\end{figure*}

We observed that, in general, the first layer filters that our CNN architecture learns are consistent with the AASM sleep scoring manual's guidelines (see Figure \ref{fig:visualization}). The first instance of consistency with the AASM sleep scoring manual are the 1-1.5 Hz and 1-1.5/12.5-14 Hz filters associated with stage N3 epochs. As shown in Table \ref{table:scoring_criteria} stage N3 is associated with activity $<$2 Hz. Interestingly, activity in 12.5-14 Hz is associated with sleep spindles, a characteristic pattern of stage N2, that can however potentially persist to stage N3 \cite[p. 27]{iber2007}. These filters exhibit little to no activation for stage N2. Therefore, it appears that the CNN learns that there are certain stage N3 epochs in which sleep spindles persist. The second instance of consistency with the AASM manual are the 13-14.5 Hz and 13-14.5/2-4 Hz filters associated with stage N2 epochs. Clearly, the 13-14.5 Hz filters are learned to capture sleep spindles, and the 13-14.5/2-4 Hz filters add K complexes (2-4 Hz activity). Interestingly, K complexes are known to be commonly followed by sleep spindles. In response to that, the CNN does not learn separate 2-4 Hz filters, but only filters that combine the detection of K complexes with the detection of subsequent sleep spindles. We think that this particular specialization of 1-1.5/12.5-14 Hz (stage N3) and 13-14.5/2-4 Hz (stage N2) filters is an indication of the power of CNNs. Incorporating such patterns into a hand-engineered features approach would demand both extensive prior knowledge as well as time-consuming filter design, while with a CNN these patterns are learned directly from the data.

The CNN also learns consistent across folds filters for stage R epochs. But while 2-5 Hz activity is clearly described in the AASM manual, these filters consistently exhibit two other characteristics: activity around 14 Hz, and \textit{absence} of activity in high frequencies (20-50 Hz). It is instructive to examine these characteristics in conjunction with the filters for stage N1 epochs, since stages N1 and R are frequently confused with one another, as shown in the confusion matrix of Table \ref{table:confusion_matrix_FpzCz}. Stage N1 is the most misclassified class, which can be also seen by the fact that filters for stage N1 are the least consistent among folds. Moreover, filters that exhibit high activation for stage N1 epochs exhibit high activation for other sleep stages as well. However, there is one stage N1 filter which appears in more than half of the folds. It has two spikes of activity around 7 Hz and around 9 Hz, and always exhibits activity in high frequencies (20-50 Hz). There is evidence in the literature that features from modalities other than EEG, such as eye movements \cite{yuval2008}, and EMG activity \cite{goncharova2003, whitham2008} can manifest themselves in the high frequencies of EEG. As shown in Tables \ref{table:scoring_criteria} and \ref{table:transition_rules}, eye movements and chin EMG tone are features that can differentiate stages N1 and R. We hypothesized that the differences between the stage R and stage N1 filters in the 20-50 Hz frequency range are related to those scoring rules. As in the case of the stage N2 and stage N3 filters described above, extensive prior knowledge and manual tweaking of filters would be required to design those filters for stages N1 and R, while with a CNN these patterns are learned directly from the data.

Finally, there are another two general characteristics in the filters that are consistent across all the folds. The first is that there are almost no filters with activity in 10.5-12 Hz. One reason that we believe this is happened is that this is the frequency region in which alpha (8-13 Hz) activity and sleep spindles (12-15 Hz) overlap, which would not be beneficial for distinguishing stages N1 and W from stage N2 (see Table \ref{table:scoring_criteria}). The second general characteristic is the absence of 15-17 Hz activity in any of the filters.

\section{Conclusion}
We showed that a CNN can achieve performance in automatic sleep stage scoring comparable to a state-of-the-art hand-engineered feature approach \cite{tsinalis2015}, without utilizing any prior knowledge from the AASM manual \cite{iber2007} that human experts follow, using a single channel of EEG (FpzCz). We analyzed and visualized the filters learned by the CNN, and discovered that the CNN learns filters that closely capture the AASM manual's guidelines in terms of their frequency characteristics per sleep stage. Our work shows that end-to-end training in CNNs is not only effective in terms of sleep stage scoring performance, but the CNN model's filters are interpretable in the context of the sleep scoring rules, and are consistent across folds in cross-validation. Outside of automatic sleep stage scoring, our work can have applications in other biosignal-based (e.g. EEG and ECG) classification problems. In particular, our analysis and visualization of the learned filters can prove useful in novel applications for which very little domain knowledge is available. For those applications, analyzing and visualizing the learned CNN filters can assist in advancing the understanding of the neurophysiological characteristics of a particular application. Using our methodology CNNs can be turned from an automation tool into a scientific tool.

\section*{Acknowledgment}
The research leading to these results was funded by the UK Engineering and Physical Sciences Research Council (EPSRC) through grant EP/K503733/1. PMM acknowledges support from the Edmond J Safra Foundation and from Lily Safra and the Imperial College Healthcare Trust Biomedical Research Centre and is an NIHR Senior Investigator. OT thanks Akara Supratak for useful discussions.

\end{document}